%% file: main.tex
\documentclass[10pt,twocolumn,letterpaper]{article}

\usepackage{wacv}
\usepackage{times}
\usepackage{epsfig}
\usepackage{graphicx}
\usepackage{amsmath}
\usepackage{amssymb}
\usepackage{booktabs}

\usepackage{algorithm2e}
\usepackage{bbm}
\usepackage{listings}
\usepackage{xcolor}
\usepackage{caption}
\usepackage{subcaption}
\usepackage[all]{nowidow}
\usepackage{amsthm}
\usepackage{enumitem}
\usepackage[accsupp]{axessibility}

\usepackage[symbol]{footmisc}

\def\wacvPaperID{168} 

\wacvalgorithmstrack   

\wacvfinalcopy 

\ifwacvfinal
\usepackage[breaklinks=true,colorlinks,bookmarks=false]{hyperref}
\else
\usepackage[pagebackref=true,breaklinks=true,colorlinks,bookmarks=false]{hyperref}
\fi

\usepackage[capitalize]{cleveref}
\crefname{section}{Sec.}{Secs.}
\Crefname{section}{Section}{Sections}
\Crefname{table}{Table}{Tables}
\crefname{table}{Tab.}{Tabs.}
\newtheorem{prop}{Proposition}
\newtheorem*{prop*}{Proposition}

\SetKwComment{Comment}{/* }{ */}
\SetKwInput{KwData}{Input}

\graphicspath{{images/}}

\definecolor{codegreen}{rgb}{0,0.6,0}
\definecolor{codegray}{rgb}{0.5,0.5,0.5}
\definecolor{codepurple}{rgb}{0.58,0,0.82}
\definecolor{backcolour}{rgb}{0.95,0.95,0.92}

\lstdefinestyle{mystyle}{
    backgroundcolor=\color{backcolour},   
    commentstyle=\color{codegreen},
    keywordstyle=\color{magenta},
    numberstyle=\tiny\color{codegray},
    stringstyle=\color{codepurple},
    basicstyle=\ttfamily\footnotesize,
    breakatwhitespace=false,         
    breaklines=true,                 
    captionpos=b,                    
    keepspaces=true,                 
    numbers=left,                    
    numbersep=5pt,                  
    showspaces=false,                
    showstringspaces=false,
    showtabs=false,                  
    tabsize=2,
    showlines=true
}

\begin{document}

\title{Similarity Contrastive Estimation for Self-Supervised Soft Contrastive Learning}

\author{
Julien DENIZE\footnote[1]{} \footnote[2]{} \and Jaonary RABARISOA\footnote[1]{} \and Astrid ORCESI\footnote[1]{} \and Romain HÉRAULT\footnote[2]{} \and Stéphane CANU\footnote[2]{} \and \\
\footnote[1]{}  Université Paris-Saclay, CEA, LIST, F-91120, Palaiseau, France \\
{\tt\small firstname.lastname@cea.fr} \\
\footnote[2]{}  Normandie Univ, INSA Rouen, LITIS, 76801, Saint Etienne du Rouvray, France \\
{\tt\small firstname.lastname@insa-rouen.fr}
}
\maketitle

\begin{abstract}
Contrastive representation learning has proven to be an effective self-supervised learning method. Most successful approaches are based on Noise Contrastive Estimation (NCE) and use different views of an instance as positives that should be contrasted with other instances, called negatives, that are considered as noise. However, several instances in a dataset are drawn from the same distribution and share underlying semantic information. A good data representation should contain relations, or semantic similarity, between the instances. Contrastive learning implicitly learns relations but considering all negatives as noise harms the quality of the learned relations. To circumvent this issue, we propose a novel formulation of contrastive learning using semantic similarity between instances called Similarity Contrastive Estimation (SCE). Our training objective is a soft contrastive learning one. Instead of hard classifying positives and negatives, we estimate from one view of a batch a continuous distribution to push or pull instances based on their semantic similarities. This target similarity distribution is sharpened to eliminate noisy relations. The model predicts for each instance, from another view, the target distribution while contrasting its positive with negatives. Experimental results show that SCE is Top-1 on the ImageNet linear evaluation protocol at 100 pretraining epochs with $72.1\%$ accuracy and is competitive with state-of-the-art algorithms by reaching $75.4\%$ for 200 epochs with multi-crop. We also show that SCE is able to generalize to several tasks. Source code is available here: \href{https://github.com/CEA-LIST/SCE}{https://github.com/CEA-LIST/SCE}.
\end{abstract}

\input{parts/introduction}
\input{parts/related_works}
\input{parts/method}
\input{parts/empirical_study/empirical_study}





\section{Conclusion}\label{Conclusion}
In this paper we introduced a self-supervised soft contrastive learning approach called Similarity Contrastive Estimation (SCE). It contrasts pairs of asymmetrical augmented views with other instances while maintaining relations among instances. The similarity distribution that defines relations is computed on one view and sharpened to remove noisy relations. SCE leverages contrastive learning and relational learning and improves the performance over optimizing only one aspect. We showed that it is competitive with the state of the art on the linear evaluation protocol on ImageNet, for fewer pretraining epochs, and to generalize to several downstream tasks. We proposed a simple but effective initial estimation of the true distribution of similarity among instances. An interesting perspective would be to propose a finer estimation of this distribution.

\section{Societal impact}
SCE as a self-supervised method for computer vision trains deep neural networks architectures that often have an economical and environmental negative impacts. But, SCE can be trained with small batches and few epochs to limit these impacts. We released our code and pretrained weights 
to limit duplicate pretraining and support the community.


\section*{Acknowledgement}\label{Acknowledgement}
This publication was made possible by the use of the Factory-AI supercomputer, financially supported by the Ile-de-France Regional Council.

\newpage

{\small
\bibliographystyle{ieee_fullname}
\bibliography{main}
}

\clearpage

\begin{appendix}
\input{parts/supplementary}

\vfill\null
\end{appendix}

\end{document}

%% file: parts/introduction.tex
\section{Introduction}
\label{sec:intro}

Self-Supervised learning (SSL) is an unsupervised learning procedure in which the data provides its own supervision to learn a practical representation of the data. It has been successfully applied to various applications such as classification and object detection. A pretext task is designed on the data to pretrain the model. The pretrained model is then fine-tuned on downstream tasks and several works have shown that a self-supervised pretrained network can outperform its supervised counterpart \cite{Caron2020, Grill2020, Caron2021}.

Contrastive learning is a state-of-the-art self-supervised paradigm based on Noise Contrastive Estimation (NCE) \cite{Gutmann2010} whose most successful applications rely on instance discrimination \cite{He2020, Chen2020b}. Pairs of views from same images are generated by carefully designed data augmentations \cite{Chen2020b, Tian2020a}. Elements from the same pairs are called \emph{positives} and their representations are pulled together to learn view invariant features. Other images called \emph{negatives} are considered as noise and their representations are pushed away from positives. Frameworks based on contrastive learning paradigm require a procedure to sample positives and negatives to learn a good data representation. A large number of negatives is essential \cite{VanDenOord2018} and various strategies have been proposed to enhance the number of negatives \cite{Chen2020b, Wu2018, He2020, Kalantidis2020}. Sampling hard negatives \cite{Kalantidis2020, Robinson2021, Wu2021, Hu2021, Dwibedi2021} improve the representations but can be harmful if they are semantically false negatives which is known as the "class collision problem" \cite{Cai2020, Wei2021, Chuang}. 

Other approaches that learn from positive views without negatives have been proposed by predicting pseudo-classes of different views \cite{Caron2020, Caron2021}, minimizing the feature distance of positives \cite{Grill2020, Chen2021a} or matching the similarity distribution between views and other instances \cite{Zheng2021}. These methods free the mentioned problem of sampling hard negatives.

Based on the weakness of contrastive learning using negatives, we introduce a self-supervised soft contrastive learning approach called Similarity Contrastive Estimation (SCE), that contrasts positive pairs with other instances and leverages the push of negatives using the inter-instance similarities. Our method computes relations defined as a sharpened similarity distribution between augmented views of a batch. Each view from the batch is paired with a differently augmented query. Our objective function will maintain for each query the relations and contrast its positive with other images. A memory buffer is maintained to produce a meaningful distribution. Experiments on several datasets show that our approach outperforms our contrastive and relational baselines MoCov2 \cite{Chen2020a} and ReSSL \cite{Zheng2021}.

Our contributions can be summarized as follows:
\begin{itemize}[noitemsep,topsep=0pt,leftmargin=*]
\item We propose a self-supervised soft contrastive learning approach called Similarity Contrastive Estimation (SCE) that contrasts pairs of augmented images with other instances and maintains relations among instances.
\item We demonstrate that our framework SCE outperforms on several benchmarks its baselines MoCov2 \cite{Chen2020a} and ReSSL \cite{Zheng2021} for a shared architecture and can further be improved using more recent architectures with a larger batch size and a predictor.
\item We show that our proposed SCE is competitive with the state of the art on the ImageNet linear evaluation protocol and generalizes to several downstream tasks.
\end{itemize}

%% file: parts/related_works.tex
\section{Related Work}\label{Related Works}

\textbf{Self-Supervised Learning.} In early works, different \emph{pretext tasks} to perform Self-Supervised Learning have been proposed to learn a good data representation such as: instance discrimination \cite{Dosovitskiy2016}, patch localization \cite{Doersch2015}, colorization \cite{Zhang2016}, jigsaw puzzle \cite{Noroozi2016}, counting \cite{Noroozi2017}, angle rotation prediction \cite{Gidaris2018}.

\textbf{Contrastive Learning.} Contrastive learning is a learning paradigm \cite{VanDenOord2018, Wu2018, DevonHjelm2018a, Tian2020, He2020, Chen2020b, Misra2020, Tian2020a, Caron2020, Grill2020} that outperformed previously mentioned \emph{pretext tasks}. Most successful methods rely on instance discrimination with a \emph{positive} pair of views from the same image contrasted with all other instances called \emph{negatives}. Retrieving lots of negatives is necessary for contrastive learning \cite{VanDenOord2018} and various strategies have been proposed. MoCo (v2) \cite{He2020, Chen2020a} uses a small batch size and keeps a high number of negatives by maintaining a memory buffer of representations via a momentum encoder. Alternatively, SimCLR \cite{Chen2020b, Chen2020d} and MoCov3 \cite{Chen2021b} use a large batch size without a memory buffer, and without a momentum encoder for SimCLR.

\textbf{Sampler for Contrastive Learning.} All negatives are not equal \cite{Cai2020} and hard negatives, negatives difficult to distinguish with positives, are the most important to sample to improve contrastive learning. However, they are potentially harmful to the training because of the ``class collision" problem \cite{Cai2020, Wei2021, Chuang}. Several samplers have been proposed to alleviate this problem such as using the nearest neighbor as positive for NNCLR \cite{Dwibedi2021}. Truncated-triplet \cite{Wang2021a} optimizes a triplet loss using the k-th similar element as negative that showed significant improvement. It is also possible to generate views by adversarial learning as AdCo \cite{Hu2021} showed.

\textbf{Contrastive Learning without negatives.} Various siamese frameworks perform contrastive learning without the use of negatives to avoid the class collision problem. BYOL \cite{Grill2020} trains an online encoder to predict the output of a momentum updated target encoder. SwAV \cite{Caron2020} enforces consistency between online cluster assignments from learned prototypes. DINO \cite{Caron2021} proposes a self-distillation paradigm to match distribution on pseudo class from an online encoder to a momentum target encoder. Barlow-Twins \cite{Zbontar2021} aligns the cross-correlation matrix between two paired outputs to the identity matrix that VICReg \cite{Bardes2022} stabilizes by adding an intra-batch decorrelation loss function.

\textbf{Regularized Contrastive Learning.} Several works regularize contrastive learning by optimizing a contrastive objective along with an objective that considers the similarities among instances. CO2 \cite{Wei2021} 
adds a consistency regularization term that matches the distribution of similarity for a query and its positive. PCL \cite{Li2021} and WCL \cite{Zheng2021b} combines unsupervised clustering with contrastive learning to tighten representations of similar instances.  

\textbf{Relational Learning.} Contrastive learning implicitly learns relations among instances by optimizing alignment and matching a prior distribution \cite{Wang2020, Chen2020c}. ReSSL \cite{Zheng2021} introduces an explicit relational learning objective by maintaining consistency of pairwise similarities between strong and weak augmented views. The pairs of views are not directly aligned which harms the discriminative performance.

In our work, we optimize a contrastive learning objective using negatives that alleviate class collision by pulling related instances. We do not use a regularization term but directly optimize a soft contrastive learning objective that leverages the contrastive and relational aspects. 

%% file: parts/method.tex
\section{Methodology}\label{Methodology}

\begin{figure*}
\begin{subfigure}[b]{0.5\textwidth}
 \centering
\includegraphics[width=0.5\textwidth]{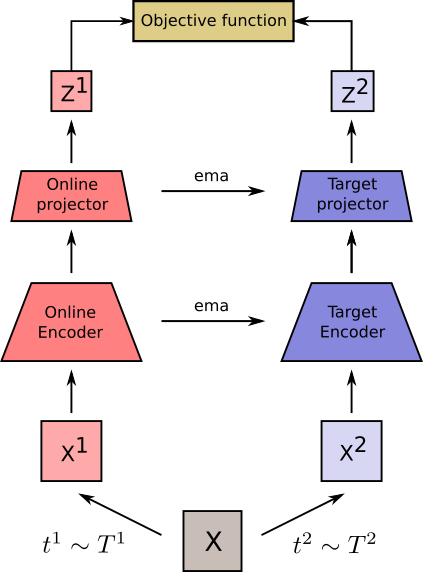}
 \caption{Siamese pipeline}
 \label{fig:siamese}
\end{subfigure}
\begin{subfigure}[b]{0.5\textwidth}
 \centering
\includegraphics[width=0.80\textwidth]{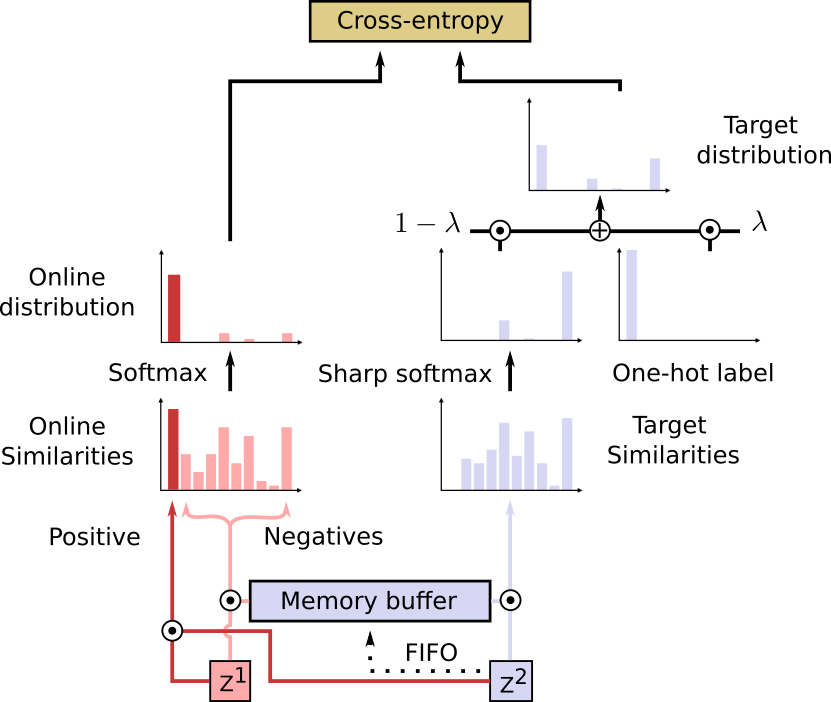}
 \caption{SCE objective function}
 \label{fig:sce}
\end{subfigure}
\caption{SCE follows a siamese pipeline illustrated in \cref{fig:siamese}. A batch $\mathbf{x}$ of images is augmented with two different data augmentation distributions $T^1$ and $T^2$  to form $\mathbf{x^1} = t^1(\mathbf{x})$ and $\mathbf{x^2} = t^2(\mathbf{x})$ with $t^1 \sim T^1$ and $t^2 \sim T^2$. The representation $\mathbf{z^1}$ is computed through an online encoder $f_s$ and projector $g_s$ such as $\mathbf{z^1} = g_s(f_s(\mathbf{x^1}))$. A parallel target branch updated by an exponential moving average of the online branch, or ema, computes $\mathbf{z^2} = g_t(f_t(\mathbf{x^2}))$ with $f_t$ and $g_t$ the target encoder and projector. In the objective function of SCE illustrated in \cref{fig:sce}, $\mathbf{z^2}$ is used to compute the inter-instance target distribution by applying a sharp softmax to the cosine similarities between $\mathbf{z^2}$ and a memory buffer of representations from the momentum branch. This distribution is mixed via a $1 - \lambda$ factor with a one-hot label factor $\lambda$ to form the target distribution. Similarities between $\mathbf{z^1}$ and the memory buffer plus its positive in $\mathbf{z^2}$ are also computed. The online distribution is computed via softmax applied to the online similarities. The objective function is the cross entropy between the target and the online distributions.}
\label{fig:pipeline}
\end{figure*}

In this section, we will introduce our baselines: MoCov2 \cite{Chen2020a} for the contrastive aspect and ReSSL \cite{Zheng2021} for the relational aspect. We will then present our self-supervised soft contrastive learning approach called Similarity Contrastive Estimation (SCE). All these methods share the same architecture illustrated in \cref{fig:siamese}. We provide the pseudo-code of our algorithm in supplementary material.

\subsection{Contrastive and Relational Learning}\label{Contrastive and Relational Learning}
Consider $\mathbf{x}=\{\mathbf{x_k}\}_{k\in \{1, ..., N\}}$ a batch of $N$ images. Siamese momentum methods based on Contrastive and Relational learning, such as MoCo \cite{He2020} and ReSSL \cite{Zheng2021} respectively, produce two views of $\mathbf{x}$, $\mathbf{x^1} = t^1(\mathbf{x})$ and $\mathbf{x^2} = t^2(\mathbf{x})$, from two data augmentation distributions $T^1$ and $T^2$ with $t^1 \sim T^1$ and $t^2 \sim T^2$.  For ReSSL, $T^2$ is a weak data augmentation distribution compared to $T^1$ to maintain relations. $\mathbf{x^1}$ passes through an online network $f_s$ followed by a projector $g_s$ to compute $\mathbf{z^1} = g_s(f_s(\mathbf{x^1}))$. A parallel target branch containing a projector $g_t$ and an encoder $f_t$ updated by exponential moving average of the online branch computes $\mathbf{z^2} = g_t(f_t(\mathbf{x^2}))$. $\mathbf{z^1}$ and $\mathbf{z^2}$ are both $l_2$-normalized.

MoCo uses the InfoNCE loss, a similarity based function scaled by the temperature $\tau$ that maximizes agreement between the positive pair and push negatives away:

\begin{equation}
  L_{InfoNCE} = - \frac{1}{N} \sum_{i=1}^N \log\left(\frac{\exp(\mathbf{z^1_i} \cdot \mathbf{z^2_i} / \tau)}{\sum_{j=1}^N\exp(\mathbf{z^1_i} \cdot \mathbf{z^2_j} / \tau)}\right).
  \label{eq:1}
\end{equation}

ReSSL computes a target similarity distribution $\mathbf{s^2}$, that represents the relations between weak augmented instances, and the distribution of similarity $\mathbf{s^{1}}$ between the strongly augmented instances with the weak augmented ones. Temperature parameters are applied to each distribution: $\tau$ for $\mathbf{s^{1}}$ and $\tau_m$ for $\mathbf{s^{2}}$ with $\tau > \tau_m$ to eliminate noisy relations. The loss function is the cross-entropy between $\mathbf{s^{2}}$ and $\mathbf{s^{1}}$:

\begin{equation}
  s^{1}_{ik} = \frac{\mathbbm{1}_{i \neq k} \cdot \exp(\mathbf{z^1_i} \cdot \mathbf{z^2_k} / \tau)}{\sum_{j=1}^{N}{\mathbbm{1}}_{i \neq j} \cdot \exp(\mathbf{z^1_i} \cdot \mathbf{z^2_j} / \tau)},
  \label{eq:2}
 \end{equation}
 
 \begin{equation}
  s^{2}_{ik} = \frac{{\mathbbm{1}}_{i \neq k} \cdot \exp(\mathbf{z^2_i} \cdot \mathbf{z^2_k} / \tau_m)}{\sum_{j=1}^{N}{\mathbbm{1}}_{i \neq j} \cdot \exp(\mathbf{z^2_i} \cdot \mathbf{z^2_j} / \tau_m)},
  \label{eq:3}
\end{equation}

\begin{equation}
  L_{ReSSL} = - \frac{1}{N} \sum_{i=1}^N\sum_{\substack{k=1 \\ k\neq i}}^N s^{2}_{ik} \log\left(s^{1}_{ik}\right).
  \label{eq:4}
\end{equation}


A memory buffer of size $M >> N$ filled by $\mathbf{z^2}$ is maintained for both methods.







\subsection{Similarity Contrastive Estimation}\label{Similarity Contrastive Estimation}


Contrastive Learning methods damage relations among instances which Relational Learning correctly build. However Relational Learning lacks the discriminating features that contrastive methods can learn. If we take the example of a dataset composed of cats and dogs, we want our model to be able to understand that two different cats share the same appearance but we also want our model to learn to distinguish details specific to each cat. Based on these requirements, we propose our approach called Similarity Contrastive Estimation (SCE).

We argue that there exists a true distribution of similarity $\mathbf{w_i^*}$ between a query $\mathbf{q_i}$ and the instances in a batch of $N$ images $\mathbf{x}=\{\mathbf{x_k}\}_{k\in \{1, ..., N\}}$, with $\mathbf{x_i}$ a positive view of $\mathbf{q_i}$. If we had access to $\mathbf{w_i^*}$, our training framework would estimate the similarity distribution $\mathbf{p_i}$ between $\mathbf{q_i}$ and all instances in $\mathbf{x}$, and minimize the cross-entropy between $\mathbf{w_i^*}$ and $\mathbf{p_i}$ which is a soft contrastive learning objective:

\begin{equation}
L_{SCE^*} = - \frac{1}{N}\sum_{i=1}^N\sum_{k=1}^N w^*_{ik}\log\left(p_{ik}\right).
\label{eq:5}
\end{equation}

$L_{SCE^*}$ is a soft contrastive approach that generalizes InfoNCE and ReSSL objectives. InfoNCE is a hard contrastive loss that estimates $\mathbf{w_i^*}$ with a one-hot label and ReSSL estimates $\mathbf{w_i^*}$ without the contrastive component. 

We propose an estimation of $\mathbf{w_i^*}$ based on contrastive and relational learning. We consider $\mathbf{x^1} = t^1(\mathbf{x})$ and $\mathbf{x^2} = t^2(\mathbf{x})$ generated from $\mathbf{x}$ using two data augmentations $t^1 \sim T^1$ and $t^2 \sim T^2$.
Both augmentation distributions should be different to estimate different relations for each view. We compute $\mathbf{z^1} = g_s(f_s(\mathbf{x^1}))$ from $f_s$ and $g_s$ (and optionally a predictor \cite{Grill2020, Chen2021b}) and $\mathbf{z^2} = g_t(f_t(\mathbf{x^2}))$. $\mathbf{z^1}$ and $\mathbf{z^2}$ are both $l_2$-normalized. The similarity distribution $\mathbf{s^2}$ that defines relations among instances is computed via the \cref{eq:3}. The temperature $\tau_m$ sharpens the distribution to only keep relevant relations. A weighted positive one-hot label is added to $\mathbf{s^2_i}$ to build the target similarity distribution $\mathbf{w^2_i}$:

\begin{equation}
  w^2_{ik} = \lambda \cdot \mathbbm{1}_{i=k} + (1 - \lambda) \cdot s^2_{ik}.
  \label{eq:7}
\end{equation}

The online similarity distribution $\mathbf{p^1_i}$ between $\mathbf{z^1_i}$ and $\mathbf{z^2}$, including $\mathbf{z^2_i}$ in opposition with ReSSL, is computed and scaled by the temperature $\tau$ with $\tau > \tau_m$ to build a sharper target distribution:

\begin{equation}
  p^1_{ik} = \frac{\exp(\mathbf{z^1_i} \cdot \mathbf{z^2_k} / \tau)}{\sum_{j=1}^{N}\exp(\mathbf{z^1_i} \cdot \mathbf{z^2_j} / \tau)}.
  \label{eq:8}
\end{equation}

The objective function illustrated in \cref{fig:sce} is the cross-entropy between each $\mathbf{w^2}$ and $\mathbf{p^1}$:
\begin{equation}
  L_{SCE} = - \frac{1}{N} \sum_{i=1}^N\sum_{k=1}^N w^2_{ik} \log\left(p^1_{ik}\right).
  \label{eq:9}
\end{equation}

The loss can be symmetrized by passing $\mathbf{x^1}$ through the momentum encoder, $\mathbf{x^2}$ through the online encoder and averaging the two losses computed.

A memory buffer of size $M >> N$ filled by $\mathbf{z^2}$ is maintained to better approximate the similarity distributions. 

The following proposition explicitly shows that SCE optimizes a contrastive learning objective while maintaining inter-instance relations:
\begin{prop}
\label{prop:1}
$L_{SCE}$ defined in \cref{eq:9} can be written as:
\begin{equation}
  L_{SCE} = \lambda \cdot L_{InfoNCE} + \mu \cdot L_{ReSSL} +  \eta \cdot L_{Ceil},
  \label{eq:10}
\end{equation}
with $\mu = \eta = 1 - \lambda$ and \newline $L_{Ceil} = - \frac{1}{N} \sum_{i=1}^{N}\log\left(\frac{\sum_{j=1}^{N}\mathbbm{1}_{i \neq j} \cdot \exp(\mathbf{z^1_i} \cdot \mathbf{z^2_j} / \tau)}{\sum_{j=1}^{N} \exp(\mathbf{z^1_i} \cdot \mathbf{z^2_j} / \tau)}\right)$.
\end{prop}

The proof separates the positive term and negatives. It can be found in the supplementary material. $L_{Ceil}$ leverages how similar the positives should be with hard negatives. Because our approach is a soft contrastive learning objective, we optimize the formulation in \cref{eq:9} and have the constraint $\mu = \eta = 1 - \lambda$. It frees our implementation from having three losses to optimize with two hyperparameters $\mu$ and $\eta$ to tune. Still, we performed a small study of the objective defined in \cref{eq:10} without this constraint to check if $L_{Ceil}$ improves results in \cref{Ablative study}.

%% file: parts/empirical_study/empirical_study.tex
\section{Empirical study}\label{Empirical study}

In this section, we first make an ablative study of our approach Similarity Contrastive Estimation (SCE) to find the best hyperparameters. Secondly, we compare SCE with its baselines MoCov2 \cite{Chen2020a} and ReSSL \cite{Zheng2021}. Finally, we evaluate SCE on the ImageNet Linear evaluation protocol and assess its generalization capacity on various tasks.

\input{parts/empirical_study/ablative_study}
\input{parts/empirical_study/baselines}
\input{parts/empirical_study/linear_classification_protocol}
\input{parts/empirical_study/transfer_learning}

%% file: parts/empirical_study/ablative_study.tex
\subsection{Ablation study}\label{Ablative study}

\begin{table*}[!htbp]
  \centering
  \begin{tabular}{c|ccccccccccc}
    \hline
    $\lambda$ & 0.    & 0.1   & 0.2   & 0.3   & 0.4   & 0.5   & 0.6   & 0.7   & 0.8   & 0.9   & 1.0 \\
    Top-1     & 81.53 & 81.77 & 82.54 & 82.81 & 82.91 & \textbf{82.94} & 82.17 & 81.58 & 81.75 & 81.79 & 81.11 \\
    \hline
  \end{tabular}
  \caption{Effect of varying $\lambda$ on the Top-1 accuracy on ImageNet100. $\lambda = 0.5$ is optimal confirming that learning to discriminate and maintaining relations is best.}
  \label{tab:lambda}
\end{table*}

\begin{table*}[!htbp]
    \centering
    \begin{tabular}{lccccc}
        \hline
        Parameter                           & \emph{weak} & \emph{strong} & \emph{strong-$\alpha$} & \emph{strong-$\beta$} & \emph{strong-$\gamma$} \\ \hline
        Random crop probability             & 1    & 1      & 1       & 1        & 1 \\ 
        Flip probability                    & 0.5  & 0.5    & 0.5     & 0.5      & 0.5 \\
        Color jittering probability         & 0.   & 0.8    & 0.8     & 0.8      & 0.8 \\
        Brightness adjustment max intensity & -    & 0.4    & 0.4     & 0.4      & 0.4 \\  
        Contrast adjustment max intensity   & -    & 0.4    & 0.4     & 0.4      & 0.4 \\
        Saturation adjustment max intensity & -    & 0.4    & 0.2     & 0.2      & 0.2 \\
        Hue adjustment max intensity        & -    & 0.1    & 0.1     & 0.1      & 0.1 \\
        Color dropping probability          & 0.   & 0.2    & 0.2     & 0.2      & 0.2 \\
        Gaussian blurring probability       & 0.   & 0.5    & 1.      & 0.1      & 0.5 \\
        Solarization probability            & 0.   & 0.     & 0.      & 0.2      & 0.2 \\ \hline    
    \end{tabular}
    \caption{Different distributions of data augmentations applied to SCE. The \emph{weak} distribution is the same as ReSSL \cite{Zheng2021}, \emph{strong} is the standard contrastive data augmentation \cite{Chen2020b}. The \emph{strong-$\alpha$} and \emph{strong-$\beta$} are two distributions introduced by BYOL \cite{Grill2020}. Finally, \emph{strong-$\gamma$} is a mix between \emph{strong-$\alpha$} and \emph{strong-$\beta$}.}
    \label{tab:augmentations}
\end{table*}


\begin{table}[!htbp]
    \centering
    \begin{tabular}{ccc|cc}
    \hline
    \multicolumn{3}{c|}{Loss coefficients} & \multicolumn{2}{c}{Top-1} \\
   $\lambda $ & $\mu$ & $\eta$ & $\tau_m = 0.05$ & $\tau_m = 0.07$ \\ \hline
   1.        & 0.    & 0.         & 81.11 & 81.11 \\
   0.5       & 0.5   & 0.         & 82.80  & 82.49 \\  
   0.5       & 0.5   & 0.5        & \textbf{82.94} & \textbf{83.37} \\
   0.        & 1.    & 0.         & 80.79 & 78.35 \\ 
   0.        & 1.    & 1.         & 81.53 & 79.64 \\ \hline
\end{tabular}
\caption{Effect of loss coefficients in \cref{eq:10} on the Top-1 accuracy on ImageNet100. $L_{Ceil}$ consistently improves performance that varies given the temperature parameters.}
\label{tab:loss}
\end{table}

\begin{table}[!htbp]
    \centering
    \begin{tabular}{cccc}
    \hline 
    Online aug & Teacher aug & Sym & top-1 \\ \hline
    \emph{strong}     & \emph{weak}        & no  & 82.94 \\
    \emph{strong-$\gamma$} & \emph{weak}   & no  & \textbf{83.00} \\
    \emph{weak}       & \emph{strong}      & no  & 73.43 \\
    \emph{strong}     & \emph{strong}      & no  & 80.54 \\
    \emph{strong-$\alpha$} & \emph{strong-$\beta$} & no  & 80.74 \\ \hline
    \emph{strong}     & \emph{weak}        & yes & 83.66 \\
    \emph{strong}     & \emph{strong}      & yes & 83.00 \\
    \emph{strong-$\alpha$} & \emph{strong-$\beta$} & yes & \textbf{84.17} \\ 
    \hline
    \end{tabular}
    \caption{Effect of using different distributions of data augmentations for the two views and of the loss symmetrization on the Top-1 accuracy on ImageNet100. Using a \emph{weak} view for the teacher without symmetry is necessary to obtain good relations. With loss symmetry, asymmetric data augmentations improve the results, with the best obtained using strong-$\alpha$ and  \emph{strong-$\beta$} augmentations.}
    \label{tab:aug}
\end{table}

\begin{table*}
\centering
    \begin{tabular}{lccccccc}
    \hline 
    Method & ImageNet & ImageNet100 & Cifar10 & Cifar100 & STL10 & Tiny-ImageNet \\ \hline
    MoCov2 \cite{Chen2020a}  & 67.5 & -  & - & - & - & - \\
    MoCov2 [*]  & 68.8 & 80.46  & 87.56 & 61.00 & 86.53 & 45.93 \\
    ReSSL \cite{Zheng2021} & 69.9 & - & 90.20  & 63.79 & 88.25 & 46.60 \\ 
    ReSSL [*] & 70.2 & 81.58 & 90.20  & 64.01 & 89.05 & 49.47 \\ 
    \textbf{SCE (Ours)}  & \textbf{70.5} & \textbf{83.37} & \textbf{90.34} & \textbf{65.45} & \textbf{89.94} & \textbf{51.90} \\ \hline
    \end{tabular}
    \caption{Comparison of SCE with its baselines MoCov2 \cite{Chen2020a} and ReSSL \cite{Zheng2021} on the Top-1 Accuracy on various datasets. SCE outperforms on all benchmarks its baselines. [*] denotes our reproduction.}
    \label{tab:baselines}
\end{table*}

\begin{table}[!htbp]
  \centering
    \begin{tabular}{cc|cc}
    \hline 
    \multicolumn{2}{c}{$\tau = 0.1$} & \multicolumn{2}{|c}{$\tau = 0.2$} \\
    $\tau_m$ & Top-1 & $\tau_m$ & Top-1 \\ \hline
    0.03 & 82.33 & 0.03 & \textbf{81.28} \\ 
    0.04 & 82.52 & 0.04 & 81.15 \\
    0.05 & 82.94 & 0.05 & 81.19\\
    0.06 & 82.54 & 0.06 & 81.19 \\ 
    0.07 & \textbf{83.37} & 0.07 & 81.13 \\ 
    0.08 & 82.71 & 0.08 & 80.91 \\ 
    0.09 & 82.53 & 0.09 & 81.18\\ 
    0.10 & 82.07 & 0.10 & 81.20 \\ \hline
    \end{tabular}
    \caption{Effect of varying the temperature parameters $\tau_m$ and $\tau$ on the Top-1 accuracy on ImageNet100. $\tau_m$ is lower than $\tau$ to produce a sharper target distribution without noisy relations. Our approach does not collapse when $\tau_m \rightarrow \tau$.}
    \label{tab:temperature}
\end{table}

To make the ablation studies, we conducted experiments on ImageNet100 that has a close distribution to ImageNet, studied in \cref{ImageNet Linear Protocol}, with the advantage to require less resources to train. We keep implementation details close to ReSSL \cite{Zheng2021} and MoCov2 \cite{Chen2020a} to ensure fair comparison.

\textbf{Dataset.} ImageNet \cite{Deng2009} is a large dataset with 1k classes, almost 1.3M images in the training set and 50K images in the validation set. ImageNet100 is a selection of 100 classes from ImageNet whose classes have been selected randomly. We took the selected classes from \cite{Tian2020} referenced in the supplementary material.

\textbf{Implementation details for pretraining.} We use the ResNet-50 \cite{He2016} encoder and pretrain for 200 epochs. As for ReSSL \cite{Zheng2021}, we apply by default \emph{strong} and \emph{weak} data augmentations defined in \cref{tab:augmentations}. We use 8 GPUs with a batch size of 512. The memory buffer size is 65,536. The projector is a 2 fully connected layer network with a hidden dimension of 4096 and an output dimension of 256. The SGD optimizer \cite{Sutskever2013} is used with a momentum of 0.9 and a weight decay of $10^{-4}$. A linear warmup is applied during 5 epochs to reach the initial learning rate of $0.3$. The learning rate is scaled using the linear scaling rule and follows the cosine decay scheduler without restart \cite{Loshchilov2016}. The momentum value to update the momentum network follows a cosine strategy from 0.996 to 1. We do not symmetrize the loss by default.

\textbf{Evaluation protocol.} To evaluate our pretrained encoders,  we train a linear classifier for 100 epochs on top of the frozen pretrained encoder using an SGD optimizer with an initial learning rate of $30$ without weight decay and a momentum of 0.9. The learning rate is decayed by a factor of $0.1$ at 60 and 80 epochs. Data augmentations follow standard protocol \cite{Chen2020a, Zheng2021}, available in supplementary material.


\textbf{Leveraging contrastive and relational learning.} 
SCE defined in \cref{eq:7} leverages contrastive and relational learning via the $\lambda$ coefficient. We studied the effect of varying $\lambda$ on ImageNet100. Temperature parameters are set to $\tau = 0.1$ and $\tau_m = 0.05$. We report the results in \cref{tab:lambda}. Performance increases with $\lambda$ from $0$ to $0.5$ after which it starts decreasing. The best $\lambda$ is $0.5$ confirming that balancing the contrastive and relational aspects provides better representation. In next experiments, we keep $\lambda = 0.5$.


We performed a small study of the optimization of \cref{eq:10} by removing $L_{ceil}$ ($\eta = 0$) to validate the relevance of our approach for $\tau = 0.1$ and $\tau_m\in\{0.05, 0.07\}$. The results are reported in \cref{tab:loss}. Adding the term $L_{ceil}$ consistently improves performance, empirically proving that our approach is better than simply adding $L_{InfoNCE}$ and $L_{ReSSL}$. This performance boost varies with temperature parameters and our best setting improves by $+0.9$ percentage points (p.p.) in comparison with adding the two losses. 


\textbf{Asymmetric data augmentations to build the similarity distributions.}
Contrastive learning approaches use strong data augmentations \cite{Chen2020b} to learn view invariant features and prevent the model to collapse. However, these strong data augmentations shift the distribution of similarities among instances that SCE uses to approximate $w_i^*$ in \cref{eq:7}. We need to carefully tune the data augmentations to estimate a relevant target similarity distribution. We listed different distributions of data augmentations in \cref{tab:augmentations}. The \emph{weak} and \emph{strong} augmentations are the same as described by ReSSL \cite{Zheng2021}. \emph{strong-$\alpha$} and \emph{strong-$\beta$} have been proposed by BYOL \cite{Grill2020}. \emph{strong-$\gamma$} combines \emph{strong-$\alpha$} and \emph{strong-$\beta$}.

We performed a study in \cref{tab:aug} on which data augmentations are needed to build a proper target distribution for the non-symmetric and symmetric settings. We report the Top-1 accuracy on Imagenet100 when varying the data augmentations applied on the online and target branches of our pipeline. For the non-symmetric setting, SCE requires the target distribution to be built from a \emph{weak} augmentation distribution that maintains consistency across instances. 

Once the loss is symmetrized, asymmetry with strong data augmentations has better performance. Indeed, using \emph{strong-$\alpha$} and \emph{strong-$\beta$} augmentations is better than using \emph{weak} and \emph{strong} augmentations, and same \emph{strong} augmentations has lower performance. We argue symmetrized SCE requires asymmetric data augmentations to produce different relations for each view to make the model learn more information. The effect of using stronger augmentations is balanced by averaging the results on both views. Symmetrizing the loss boosts the performance as for \cite{Grill2020, Chen2021a}.



\textbf{Sharpening the similarity distributions.} 
The temperature parameters sharpen the distributions of similarity exponentially. SCE uses the temperatures $\tau_m$ and $\tau$ for the target and online similarity distributions with $\tau_m < \tau$ to guide the online encoder with a sharper distribution. We made a temperature search on ImageNet100 by varying $\tau$ in $\{0.1, 0.2\}$ and $\tau_m$ in $\{0.03, ..., 0.10\}$. The results are in \cref{tab:temperature}. We found the best values $\tau_m = 0.07$ and $\tau = 0.1$ proving SCE needs a sharper target distribution. In supplementary material, this parameter search is done for other datasets used in comparison with our baselines. Unlike ReSSL \cite{Zheng2021}, SCE does not collapse when $\tau_m \rightarrow \tau$ thanks to the contrastive aspect. Hence, it is less sensitive to the temperature choice.



%% file: parts/empirical_study/baselines.tex
\subsection{Comparison with our baselines}\label{Baselines}

We compared on 6 datasets how SCE performs against its baselines. We keep similar implementation details to ReSSL \cite{Zheng2021} and MoCov2 \cite{Chen2020a} for fair comparison.

\textbf{Small datasets.} Cifar10 and Cifar100 \cite{Krizhevsky2009} have 50K training images, 10K test images, $32 \times 32$ resolution and 10-100 classes respectively. \textbf{Medium datasets.} STL10 \cite{Coates2011} has a $96 \times 96$ resolution, 10 classes, 100K unlabeled data, 5k labeled training images and 8K test images. Tiny-Imagenet \cite{Abai2020} is a subset of ImageNet with $64 \times 64$ resolution, 200 classes, 100k training images and 10K validation images.

\textbf{Implementation details.} Implementation details for small and medium datasets are in the supplementary material. For ImageNet, we follow the ones in ablation study with some modifications. The initial learning rate is set to $0.5$, the projector is a 3 fully connected layer network with a hidden dimension of 2048, a batch normalization \cite{Ioffe2015} at each layer and an output dimension of 256. For MoCov2, the temperature used is $\tau = 0.2$ and for ReSSL we use the best temperatures reported \cite{Zheng2021}. For SCE, we use the best temperature parameters from ablation study for ImageNet and ImageNet100, and for other datasets, the best ones from supplementary material. We use the same architecture for all methods except that we use the same projector as on ImageNet100 on ImageNet for MoCov2 to improve the result.

\textbf{Evaluation protocol.} The evaluation protocol is the same as defined in the ablation study for all datasets.

Results are reported in \cref{tab:baselines}. Our baselines reproduction is validated as results are better than those reported by the authors. SCE outperforms its baselines on all datasets proving that our method is more efficient to learn discriminating features on the pretrained dataset. We observe that our approach outperforms more significantly ReSSL on smaller datasets than ImageNet, suggesting that it is more important to learn to discriminate among instances for these datasets. SCE has promising applications to domains with few data such as in medical applications. 

%% file: parts/empirical_study/linear_classification_protocol.tex
\subsection{ImageNet Linear Evaluation Protocol}\label{ImageNet Linear Protocol}

We compare SCE on the widely used ImageNet linear evaluation protocol with the state of the art. We scaled our method to a larger batch size and a deeper architecture using a predictor to match the state of the art results \cite{Grill2020, Chen2021b}.

\textbf{Implementation details.} We use the ResNet-50 \cite{He2016} encoder and apply \emph{strong-$\alpha$} and \emph{strong-$\beta$} augmentations defined in \cref{tab:augmentations} with a batch size of 4096 and a memory buffer of size 65,536. 
We follow the same training hyperparameters as \cite{Chen2021b} for the architecture. Specifically, we use the same projector and predictor, the LARS optimizer \cite{Ginsburg2018} with a weight decay of $1.5\cdot10^{-6}$ for 1000 epochs of training and $10^{-6}$ for fewer epochs. Bias and batch normalization parameters are excluded. The initial learning rate is $0.5$ for 100 epochs and $0.3$ for more epochs. It is linearly scaled for 10 epochs and it follows the cosine annealed scheduler. The momentum value follows a cosine scheduler from 0.996 for 1000 epochs, 0.99 for fewer epochs, to 1. The loss is symmetrized. For SCE specific hyperparameters, we keep the best from ablation study: $\lambda = 0.5$, $\tau = 0.1$ and $\tau_m = 0.07$.

\textbf{Multi-crop setting.} We follow \cite{Hu2021} setting and sample 6 different views. The first two views are global views as without multi-crop. The 4 local crops have a resolution of $192 \times 192$, $160 \times 160$, $128 \times 128$, $96 \times 96$ and scales ($0.172$, $0.86$), ($0.143$, $0.715$), ($0.114$, $0.571$), ($0.086$, $0.429$) on which we apply the \emph{strong-$\gamma$} data augmentation.

\textbf{Evaluation protocol.} We train a linear classifier for 90 epochs on top of the frozen encoder with a batch size of 1024 and a SGD optimizer with a momentum of 0.9. The initial learning rate is $0.1$ linearly scaled and follows a cosine annealed scheduler.


\begin{table}
    \centering
    \small
    \begin{tabular}{l|cccc}
    \hline
        Method     & 100  & 200  & 300  & 800-1000 \\ \hline
        SimCLR \cite{Chen2020b}    & 66.5 & 68.3 & -     & 70.4 \\
        MoCov2 \cite{Chen2021a}     & 67.4 & 69.9 & -     & 72.2 \\
        SwaV \cite{Caron2020}      & 66.5 & 69.1 & -     & 71.8 \\
        BYOL \cite{Grill2020}      & 66.5 & 70.6 & 72.5  & 74.3 \\
        Barlow-Twins\cite{Zbontar2021} & - & - & 71.4 & 73.2 \\
        AdCo \cite{Hu2021}         & -    & 68.6  & - & 72.8 \\
        ReSSL \cite{Zheng2021}     & -    & 71.4 & -     & -    \\
        WCL \cite{Zheng2021b}       & 68.1 & 70.3 & -     & 72.2 \\
        VICReg \cite{Bardes2022}  & - & - & - & 73.2 \\
        UniGrad \cite{Tao2022}   & \underline{70.3} & -    & -     & -    \\
        MoCov3 \cite{Chen2021b}    & 68.9 & -    & \underline{72.8}  & 74.6 \\
        NNCLR \cite{Dwibedi2021}     & 69.4 & 70.7 & -     & \underline{75.4} \\
        Truncated-Triplet \cite{Wang2021a} & - & \textbf{73.8} & - & \textbf{75.9} \\ \hline
        \textbf{SCE (Ours)} & \textbf{72.1} & \underline{72.7} & \textbf{73.3}  & 74.1 \\ \hline
    \end{tabular}
    \caption{State-of-the-art results on the Top-1 Accuracy on ImageNet under the linear evaluation protocol at different pretraining epochs: 100, 200, 300, 800+. SCE is Top-1 at 100 epochs and Top-2 for 200 and 300 epochs. For 800+ epochs, SCE has lower performance than several state-of the-art methods. Results style: \textbf{best}, \underline{second best}.}
    \label{tab:sota}
\end{table}

\begin{table}
\centering
    \begin{tabular}{lccc}
    \hline 
    Method      & Epochs & Top-1 \\ \hline
    UniGrad \cite{Tao2022} & 100 & 71.7\\
    UniGrad (+ Cut-Mix) \cite{Tao2022} & 100 & 72.3 \\
    SwaV \cite{Caron2020}       & 200    & 72.7 \\
    AdCo \cite{Hu2021} & 200 & 73.2 \\
    WCL \cite{Zheng2021b}        & 200    & 73.3 \\
    Truncated-Triplet \cite{Wang2021a} & 200    & 74.1 \\
    ReSSL \cite{Zheng2021}      & 200    & 74.7 \\
    WCL \cite{Zheng2021b}         & 800    & 74.7 \\
    SwaV \cite{Caron2020}       & 800    & 75.3 \\
    DINO \cite{Caron2021}        & 800    & 75.3 \\
    UniGrad (+ Cut-Mix) \cite{Tao2022} & 800 & 75.5 \\
    NNCLR \cite{Dwibedi2021}      & 1000   & 75.6 \\
    AdCo \cite{Hu2021} & 800 & 75.7 \\ \hline
    \textbf{SCE (ours)}  & 200    & 75.4 \\ \hline
    \end{tabular}
\caption{State-of-the-art results on the Top-1 Accuracy on ImageNet under the linear evaluation protocol with multi-crop. SCE is competitive with the best state-of-the-art methods by pretraining for only 200 epochs instead of 800+.}
\label{tab:sota-multi-crop}
\end{table}

We evaluated SCE at epochs 100, 200, 300 and 1000 on the Top-1 accuracy on ImageNet to study the efficiency of our approach and compare it with the state of the art in \cref{tab:sota}. At 100 epochs, SCE reaches 72.1\% up to 74.1\% at 1000 epochs. Hence, SCE has a fast convergence and few epochs of training already provides a good representation. SCE is the Top-1 method at 100 epochs and is second best for 200 and 300 epochs proving the good quality of its representation for few epochs of pretraining. 

At 1000 epochs, SCE is below several state-of-the art results. We argue that SCE suffers from maintaining a $\lambda$ coefficient to $0.5$ and that relational or contrastive aspects do not have the same impact at the beginning and at the end of pretraining. A potential improvement would be using a scheduler on $\lambda$ that varies over time.

We added multi-crop to SCE for 200 epochs of pretraining. It enhances the results but it is costly in terms of time and memory. It improves the results from 72.7\% to our best result 75.4\% (+2.7 p.p.). Therefore, SCE learns from having local views and they should maintain relations to learn better representations. We compared SCE with state-of-the-art methods using multi-crop in \cref{tab:sota-multi-crop}. SCE is competitive with top state-of-the-art methods that trained for 800+ epochs by having slightly lower accuracy than the best method using multi-crop ($-0.3$ p.p) and without multi-crop ($-0.5$ p.p). SCE is more efficient than other methods, as it reaches state-of-the-art results for fewer pretraining epochs.


%% file: parts/empirical_study/transfer_learning.tex
\begin{table*}[ht]
    \centering
    \footnotesize
    \begin{tabular}{lccccccccccc|c} 
    \hline
    Method   & Food101	& CIFAR10 &	CIFAR100 &	SUN397 & Cars & Aircraft &	VOC2007 &	DTD &	Pets &	Caltech101 &	Flowers & Avg.           \\ \hline
    SimCLR \cite{Chen2020b}    & 72.8 &	90.5 & 74.4 & 60.6 & 49.3 &	49.8 & 81.4 & 75.7 & 84.6 &	89.3 & 92.6 & 74.6 \\
    BYOL \cite{Grill2020}      & 75.3 & 91.3 & 78.4 & 62.2 & \textbf{67.8} & 60.6 & 82.5 & 75.5 & 90.4 &	94.2 & \textbf{96.1} & 79.5 \\
    NNCLR \cite{Dwibedi2021}     & 76.7 & 93.7 & 79.0 & 62.5 & 67.1 &	\textbf{64.1} & 83.0 & 75.5 & \textbf{91.8} &	91.3 & 95.1 & 80 \\
    
    \textbf{SCE (Ours)} & \textbf{77.7} & \textbf{94.8} & \textbf{80.4} & \textbf{65.3} & 65.7 &	59.6 & \textbf{84.0} & \textbf{77.1} & 90.9 & 92.7 & \textbf{96.1} & \textbf{80.4} \\ \hline
    Supervised & 72.3 & 93.6 & 78.3 & 61.9 & 66.7 &	61.0 & 82.8 & 74.9 & 91.5 &	\textbf{94.5} & 94.7 & 79.3 \\ \hline
    
    \end{tabular}
    \caption{Linear classifier trained on popular many-shot recognition datasets. SCE is Top-1 on 7 datasets and in average.}
    \label{tab:many-shot}
\end{table*}

\begin{table}
\centering
\small
    \begin{tabular}{lcccc} \hline
    Method & K = 16 & K = 32 & K = 64 & full \\\hline
    MoCov2 \cite{Chen2020a} & 76.14 & 79.16 & 81.52 & 84.60 \\
    PCLv2 \cite{Li2021} & 78.34 & 80.72 & 82.67 & 85.43 \\
    ReSSL \cite{Zheng2021} & 79.17 & 81.96 & 83.81 & 86.31 \\
    SwAV \cite{Caron2020} & 78.38 & 81.86 & 84.40 & 87.47 \\
    WCL \cite{Zheng2021b} & \textbf{80.24} & 82.97 & 85.01 & 87.75 \\
    \textbf{SCE (Ours)} & 79.47 & \textbf{83.05} & \textbf{85.47} & \textbf{88.24} \\ \hline
    \end{tabular}
    \caption{Transfer learning on low-shot image classification on Pascal VOC2007 \cite{Everingham2010}. All methods have been pretrained for 200 epochs. SCE is Top-1 when using 32-64 or all images per class and is second for 16 images per class.}
    \label{tab:low-shot-voc}
\end{table}

\begin{table}[h]
\centering
    \begin{tabular}{lcc}
    \hline
    Method            & $AP^{\emph{Box}}$   & $AP^{\emph{Mask}}$ \\ \hline
    Random            & 35.6             & 31.4 \\
    Relative-Loc \cite{Doersch2015}     & 40.0             & 35.0 \\
    Rotation-Pred \cite{Gidaris2018}    & 40.0             & 34.9 \\
    NPID \cite{Wu2018}              & 39.4             & 34.5 \\
    MoCo \cite{He2020}              & 40.9             & 35.5 \\
    MoCov2 \cite{Chen2020a}           & 40.9             & 35.5 \\
    SimCLR \cite{Chen2020b}           & 39.6             & 34.6 \\
    BYOL \cite{Grill2020}             & 40.3             & 35.1 \\
    \textbf{SCE (Ours)} & \underline{41.6} & \underline{36.0} \\
    Truncated-Triplet \cite{Wang2021a} & \textbf{41.7}    & \textbf{36.2} \\ \hline
    Supervised        & 40.0             & 34.7 \\ \hline
    \end{tabular}
    \caption{Object detection and Instance Segmentation on COCO \cite{Lin2014} training a Mask R-CNN \cite{He2017}. SCE is Top-2 on both tasks, slightly below Truncated-Triplet \cite{Wang2021a} and better than supervised training. Results style: \textbf{best}, \underline{second best}.}
    \label{tab:object-detection}
\end{table}

\subsection{Transfer Learning}
We study the generalization of our proposed SCE on several tasks using our best checkpoint obtained on ImageNet, the multi-crop setting for 200 pretrained epochs.

\textbf{Low-shot evaluation.} Low-shot transferability of our backbone is evaluated on Pascal VOC2007. We followed the protocol proposed by \cite{Zheng2021}. We select 16, 32, 64 or all images per class to train the classifier. Our results are compared with other state-of-the-art methods pretrained for 200 epochs in \cref{tab:low-shot-voc}. SCE is Top-1 for 32, 64 and all images per class and is second for 16 images per class, proving the generalization of our approach to few-shot learning.

\textbf{Linear classifier for many-shot recognition datasets.} We follow the same protocol as \cite{Grill2020, Ericsson2021} to study many-shot recognition in transfer learning on the datasets FGVC Aircraft \cite{Maji2013}, Caltech-101 \cite{Fei2004}, Standford Cars \cite{Krause2013}, CIFAR-10 \cite{Krizhevsky2009}, CIFAR-100 \cite{Krizhevsky2009}, DTD \cite{Cimpoi2014}, Oxford 102 Flowers \cite{Nilsback2008}, Food-101 \cite{Bossard2014}, Oxford-IIIT Pets \cite{Parkhi2012}, SUN397 \cite{Xiao2010} and Pascal VOC2007 \cite{Everingham2010}. These datasets cover a large variety of number of training images (2k-75k) and number of classes (10-397). For all datasets we study the Top-1 classification accuracy except for Aircraft, Caltech-101, Pets and Flowers for which we report the mean per-class accuracy and the 11-point MAP for VOC2007.

We report the performance of SCE in comparison with state-of-the-art methods in \cref{tab:many-shot}. SCE outperforms on 7 datasets all approaches. In average, SCE is above all state-of-the-art methods as well as the supervised baseline, meaning SCE is able to generalize to a wide range of datasets.

\textbf{Object detection and instance segmentation.} 
We performed object detection and instance segmentation on the COCO dataset \cite{Lin2014}. We used the pretrained network to initialize a Mask R-CNN \cite{He2017} until the C4 layer. We follow the protocol proposed by \cite{Wang2021a} and report the Average Precision for detection $AP^{Box}$ and instance segmentation $AP^{Mask}$.

We report our result in \cref{tab:object-detection} and observe that SCE is the second best method after Truncated-Triplet \cite{Wang2021a} on both metrics, by being slightly below their reported results and above the supervised setting. Therefore our proposed SCE is able to generalize to object detection and instance segmentation task beyond what the supervised pretraining can ($+1.6$ p.p. of $AP^{Box}$ and $+1.3$ p.p. of $AP^{Mask})$.

%% file: parts/supplementary.tex
\section{Pseudo-Code of SCE}

\begin{lstlisting}[language=Python, caption=Pseudo-Code of SCE in a pytorch style, basicstyle=\scriptsize]
# dataloader: loader of batches of size bsz
# epochs: number of epochs
# T1: weak distribution of data augmentations
# T2: strong distribution of data augmentations
# f1, g1: online encoder and projector
# f2, g2: momentum encoder and projector 
# queue: memory buffer
# tau: online temperature
# tau_m: momentum temperature
# lambda_: coefficient between contrastive and relational aspects

for i in range(epochs):
  for x in dataloader:
    x1, x2 = T1(x), T2(x)
    z1, z2 = g1(f1(x1)), g2(f2(x2))
    
    stop_grad(z2)
    
    sim2_pos = zeros(bsz)
    sim2_neg = einsum("nc,kc->nk", z2, queue)
    sim2 = cat([sim2_pos, sim2_neg]) / tau_m
    s2 = softmax(sim2)
    w2 = lambda_ * one_hot(sim2_pos, bsz+1) + (1 - lambda_) * s2

    sim1_pos = einsum("nc,nc->n", z1, z2)
    sim1_neg = einsum("nc,kc->nk", z1, queue)
    sim1 = cat([sim1_pos, sim1_neg]) / tau
    p1 = softmax(sim1)
    
    loss = cross_entropy(p1, w2)
    loss.backward()
    
    update(f1.params)
    update(g1.params)
    momentum_update(f2.params, f1.params)
    momentum_update(g2.params, g1.params)
    fifo_update(queue, z2)
\end{lstlisting}

\section{Proof Proposition 1. in Sec. 3.2}

\begin{prop*}
$L_{SCE}$ defined as

$$L_{SCE} = - \frac{1}{N} \sum_{i=1}^N\sum_{k=1}^N w^2_{ik} \log\left(p^1_{ik}\right),$$

can be written as:
\begin{equation*}
  L_{SCE} = \lambda \cdot L_{InfoNCE} + \mu \cdot L_{ReSSL} +  \eta \cdot L_{ceil},
\end{equation*}
with $\mu = \eta = 1 - \lambda$ and
\begin{equation*}
  L_{Ceil} = - \frac{1}{N} \sum_{i=1}^{N}\log\left(\frac{\sum_{j=1}^{N}\mathbbm{1}_{i \neq j} \cdot \exp(\mathbf{z^1_i} \cdot \mathbf{z^2_j} / \tau)}{\sum_{j=1}^{N} \exp(\mathbf{z^1_i} \cdot \mathbf{z^2_j} / \tau)}\right).
\end{equation*}
\end{prop*}

\begin{proof}
\normalsize
Recall that:
\begin{equation*}
\setlength{\abovedisplayskip}{12pt}
\setlength{\belowdisplayskip}{12pt}
\begin{aligned}
    p^1_{ik} &= \frac{\exp(\mathbf{z^1_i} \cdot \mathbf{z^2_k} / \tau)}{\sum_{j=1}^{N}\exp(\mathbf{z^1_i} \cdot \mathbf{z^2_j} / \tau)}, \\
    s^2_{ik} &= \frac{\mathbbm{1}_{i \neq k} \cdot \exp(\mathbf{z^2_i} \cdot \mathbf{z^2_k} / \tau_m)}{\sum_{j=1}^{N}\mathbbm{1}_{i \neq j} \cdot \exp(\mathbf{z^2_i} \cdot \mathbf{z^2_j} / \tau_m)}, \\
    w^2_{ik} &= \lambda \cdot \mathbbm{1}_{i=k} + (1 - \lambda) \cdot s^2_{ik}.
\end{aligned}
\end{equation*}
\normalsize
We decompose the second loss over $k$ in the definition of $L_{SCE}$ to make the proof:
\scriptsize
\begin{equation*}
\setlength{\abovedisplayskip}{12pt}
\setlength{\belowdisplayskip}{12pt}
\begin{aligned}
    L_{SCE} = & - \frac{1}{N} \sum_{i=1}^N\sum_{k=1}^N w^2_{ik} \log\left(p^1_{ik}\right) \\
            = & - \frac{1}{N} \sum_{i=1}^N \left[ w^2_{ii} \log\left(p^1_{ii}\right) + \sum_{\substack{k=1 \\ k\neq i}}^N w^2_{ik} log\left(p^1_{ik}\right)\right] \\
            = & \underbrace{- \frac{1}{N} \sum_{i=1}^N w^2_{ii} \left(p^1_{ii}\right)}_{(1)} \underbrace{- \frac{1}{N} \sum_{i=1}^N\sum_{\substack{k=1 \\ k\neq i}}^N w^2_{ik} \log\left(p^1_{ik}\right)}_{(2)}. \\
\end{aligned}
\end{equation*}
\normalsize
First we rewrite $(1)$ to retrieve the $L_{InfoNCE}$ loss.
\scriptsize
\begin{equation*}
\setlength{\abovedisplayskip}{12pt}
\setlength{\belowdisplayskip}{12pt}
\begin{aligned}
    (1) = & - \frac{1}{N} \sum_{i=1}^N w^2_{ii} log\left(p^1_{ii}\right) \\
        = & - \frac{1}{N} \sum_{i=1}^N \lambda \cdot \log\left(p^1_{ii}\right) \\
        = & - \lambda \cdot \frac{1}{N} \sum_{i=1}^N \log\left(\frac{\exp(\mathbf{z^1_i} \cdot \mathbf{z^2_i} / \tau)}{\sum_{j=1}^{N}\exp(\mathbf{z^1_i} \cdot \mathbf{z^2_j} / \tau)}\right) \\
        = & \lambda \cdot L_{InfoNCE}.
\end{aligned}
\end{equation*}
\normalsize
Now we rewrite $(2)$ to retrieve the $L_{ReSSL}$ and $L_{Ceil}$ losses.
\scriptsize
\begin{equation*}
\setlength{\abovedisplayskip}{12pt}
\setlength{\belowdisplayskip}{0pt}
\begin{aligned}
    (2) = & - \frac{1}{N} \sum_{i=1}^N\sum_{\substack{k=1 \\ k\neq i}}^N w^2_{ik} \log\left(p^1_{ik}\right) \\
        = & - \frac{1}{N} \sum_{i=1}^N\sum_{\substack{k=1 \\ k\neq i}}^N (1 - \lambda) \cdot s^2_{ik} \cdot \log\left(p^1_{ik}\right) \\
        = & - (1 - \lambda) \cdot \frac{1}{N} \sum_{i=1}^N\sum_{k=1}^N s^2_{ik} \cdot \log\left(p^1_{ik}\right) \\
        = & - (1 - \lambda) \cdot \frac{1}{N} \sum_{i=1}^N\sum_{k=1}^N \left[ s^2_{ik} \cdot \log\left(\frac{\exp(\mathbf{z^1_i} \cdot \mathbf{z^2_k} / \tau)}{\sum_{j=1}^{N}\exp(\mathbf{z^1_i} \cdot \mathbf{z^2_j} / \tau)}\right) \right]\\
        = & - (1 - \lambda) \cdot \frac{1}{N} \sum_{i=1}^N\sum_{k=1}^N \Bigg[ s^2_{ik} \cdot \Bigg( \\
          & \phantom{- (1 - \lambda) \cdot} \log\left(\exp(\mathbf{z^1_i} \cdot \mathbf{z^2_k} / \tau)\right) - \log\left(\sum_{j=1}^{N}\exp(\mathbf{z^1_i} \cdot \mathbf{z^2_j} / \tau)\right)\Bigg)\Bigg] \\
        = & - (1 - \lambda) \cdot \frac{1}{N} \sum_{i=1}^N\sum_{k=1}^N \Bigg[ s^2_{ik} \cdot \Bigg( \\
          & \phantom{- (1 - \lambda) \cdot} \log\left(\exp(\mathbf{z^1_i} \cdot \mathbf{z^2_k} / \tau)\right) - \log\left(\sum_{j=1}^{N}\exp(\mathbf{z^1_i} \cdot \mathbf{z^2_j} / \tau)\right) + \\
          & \phantom{- (1 - \lambda) \cdot} \log\left(\sum_{j=1}^{N} \mathbbm{1}_{i \neq j} \cdot \exp(\mathbf{z^1_i} \cdot \mathbf{z^2_j} / \tau)\right) - \\
          & \phantom{- (1 - \lambda) \cdot} \log\left(\sum_{j=1}^{N} \mathbbm{1}_{i \neq j} \cdot \exp(\mathbf{z^1_i} \cdot \mathbf{z^2_j} / \tau)\right) \Bigg)\Bigg] \\
\end{aligned}
\end{equation*}
\begin{equation*}
\setlength{\abovedisplayskip}{0pt}
\setlength{\belowdisplayskip}{12pt}
\begin{aligned}
        = & - (1 - \lambda) \cdot \frac{1}{N} \sum_{i=1}^N\sum_{k=1}^N \Bigg[ s^2_{ik} \cdot \Bigg( \log\left(\frac{\exp(\mathbf{z^1_i} \cdot \mathbf{z^2_k} / \tau)}{\sum_{j=1}^{N} \mathbbm{1}_{i \neq j} \cdot \exp(\mathbf{z^1_i} \cdot \mathbf{z^2_j} / \tau)}\right) + \\
          & \phantom{- (1 - \lambda) \cdot \frac{1}{N} \sum_{i=1}^N\sum_{k=1}^N \Bigg[ s^2_{ik} \cdot \Bigg(} \log\left(\frac{\sum_{j=1}^{N} \mathbbm{1}_{i \neq j} \cdot \exp(\mathbf{z^1_i} \cdot \mathbf{z^2_j} / \tau)}{\sum_{j=1}^{N}\exp(\mathbf{z^1_i} \cdot \mathbf{z^2_j} / \tau)}\right)\Bigg)\Bigg] \\
        = & - (1 - \lambda) \cdot \frac{1}{N} \sum_{i=1}^N\sum_{k=1}^N \Bigg[ s^2_{ik} \cdot \log\left(\frac{\exp(\mathbf{z^1_i} \cdot \mathbf{z^2_k} / \tau)}{\sum_{j=1}^{N} \mathbbm{1}_{i \neq j} \cdot \exp(\mathbf{z^1_i} \cdot \mathbf{z^2_j} / \tau)}\right)\Bigg] - \\
          & \phantom{-} (1 - \lambda) \cdot \frac{1}{N} \sum_{i=1}^N\sum_{k=1}^N \Bigg[ s^2_{ik} \cdot \log\left(\frac{\sum_{j=1}^{N} \mathbbm{1}_{i \neq j} \cdot \exp(\mathbf{z^1_i} \cdot \mathbf{z^2_j} / \tau)}{\sum_{j=1}^{N}\exp(\mathbf{z^1_i} \cdot \mathbf{z^2_j} / \tau)}\right)\Bigg]. \\
\end{aligned}
\end{equation*}
\normalsize
Because $s^2_{ii} = 0$ and $\mathbf{s^2_i}$ is a probability distribution, we have:
\begin{equation*}
\scriptsize
\setlength{\abovedisplayskip}{12pt}
\setlength{\belowdisplayskip}{12pt}
\begin{aligned}
    & \sum_{k=1}^N s^2_{ik} \cdot \log\left(\frac{\exp(\mathbf{z^1_i} \cdot \mathbf{z^2_k} / \tau)}{\sum_{j=1}^{N} \mathbbm{1}_{i \neq j} \cdot \exp(\mathbf{z^1_i} \cdot \mathbf{z^2_j} / \tau)}\right) = \\
    & \phantom{sum_{k=1}^N s^2_{ik}} \sum_{\substack{k=1 \\ k\neq i}}^N s^2_{ik} \cdot \log\left(\frac{\mathbbm{1}_{i \neq k} \cdot \exp(\mathbf{z^1_i} \cdot \mathbf{z^2_k} / \tau)}{\sum_{j=1}^{N} \mathbbm{1}_{i \neq j} \cdot \exp(\mathbf{z^1_i} \cdot \mathbf{z^2_j} / \tau)}\right), \\
    & \sum_{k=1}^N s^2_{ik} \cdot \log\left(\frac{\sum_{j=1}^{N} \mathbbm{1}_{i \neq j} \cdot \exp(\mathbf{z^1_i} \cdot \mathbf{z^2_j} / \tau)}{\sum_{j=1}^{N}\exp(\mathbf{z^1_i} \cdot \mathbf{z^2_j} / \tau)}\right) = \\
    & \phantom{sum_{k=1}^N s^2_{ik}} \log\left(\frac{\sum_{j=1}^{N} \mathbbm{1}_{i \neq j} \cdot \exp(\mathbf{z^1_i} \cdot \mathbf{z^2_j} / \tau)}{\sum_{j=1}^{N}\exp(\mathbf{z^1_i} \cdot \mathbf{z^2_j} / \tau)}\right) .
\end{aligned}
\end{equation*}
\normalsize
Then:
\begin{equation*}
\scriptsize
\begin{aligned}
        (2) = & - (1 - \lambda) \cdot \\
        & \frac{1}{N} \sum_{i=1}^N\sum_{\substack{k=1 \\ k\neq i}}^N \Bigg[ s^2_{ik} \cdot \log\left(\frac{\mathbbm{1}_{i \neq k} \cdot \exp(\mathbf{z^1_i} \cdot \mathbf{z^2_k} / \tau)}{\sum_{j=1}^{N} \mathbbm{1}_{i \neq j} \cdot \exp(\mathbf{z^1_i} \cdot \mathbf{z^2_j} / \tau)}\right)\Bigg] - \\
        & (1 - \lambda) \cdot \frac{1}{N} \sum_{i=1}^N \Bigg[ \log\left(\frac{\sum_{j=1}^{N} \mathbbm{1}_{i \neq j} \cdot \exp(\mathbf{z^1_i} \cdot \mathbf{z^2_j} / \tau)}{\sum_{j=1}^{N}\exp(\mathbf{z^1_i} \cdot \mathbf{z^2_j} / \tau)}\right) \Bigg] \\
        = & (1 - \lambda) \cdot L_{ReSSL} + (1 - \lambda) \cdot L_{Ceil}.
\end{aligned}
\end{equation*}
\normalsize
\end{proof}

\section{Classes to construct ImageNet100}
To build the ImageNet100 dataset, we used the classes shared by the CMC \cite{Tian2020} authors in the supplementary material of their publication. We also share these classes in \cref{tab:classes}.
\begin{table}[h]
\centering
\begin{tabular}{cccc}\hline
\multicolumn{4}{c}{100 selected classes from ImageNet} \\ \hline
n02869837 & n01749939 & n02488291 & n02107142 \\
n13037406 & n02091831 & n04517823 & n04589890 \\
n03062245 & n01773797 & n01735189 & n07831146 \\
n07753275 & n03085013 & n04485082 & n02105505 \\
n01983481 & n02788148 & n03530642 & n04435653 \\
n02086910 & n02859443 & n13040303 & n03594734 \\
n02085620 & n02099849 & n01558993 & n04493381 \\
n02109047 & n04111531 & n02877765 & n04429376 \\
n02009229 & n01978455 & n02106550 & n01820546 \\
n01692333 & n07714571 & n02974003 & n02114855 \\
n03785016 & n03764736 & n03775546 & n02087046 \\
n07836838 & n04099969 & n04592741 & n03891251 \\
n02701002 & n03379051 & n02259212 & n07715103 \\
n03947888 & n04026417 & n02326432 & n03637318 \\
n01980166 & n02113799 & n02086240 & n03903868 \\
n02483362 & n04127249 & n02089973 & n03017168 \\
n02093428 & n02804414 & n02396427 & n04418357 \\
n02172182 & n01729322 & n02113978 & n03787032 \\
n02089867 & n02119022 & n03777754 & n04238763 \\
n02231487 & n03032252 & n02138441 & n02104029 \\
n03837869 & n03494278 & n04136333 & n03794056 \\
n03492542 & n02018207 & n04067472 & n03930630 \\
n03584829 & n02123045 & n04229816 & n02100583 \\
n03642806 & n04336792 & n03259280 & n02116738 \\
n02108089 & n03424325 & n01855672 & n02090622 \\
\hline
\end{tabular}
\caption{The 100 classes selected from ImageNet to construct ImageNet100.}
\label{tab:classes}
\end{table}

\begin{table*}
    \centering
    \small
    \begin{tabular}{cccccccccc} \hline
        Dataset & $\tau$ & $\tau_m = 0.03$ & $\tau_m = 0.04$ & $\tau_m = 0.05$ & $\tau_m = 0.06$ & $\tau_m = 0.07$ & $\tau_m = 0.08$ & $\tau_m = 0.09$ & $\tau_m = 0.1$  \\ \hline
        CIFAR10 & 0.1 & 89.93 & 90.03 & 90.06 & 90.20 & 90.16 & 90.06 & 89.67 & 88.97 \\
        CIFAR10 & 0.2 & 89.98 & 90.12 & 90.12 & 90.05 & 90.13 & 90.09 & 90.22 & \textbf{90.34} \\ \hline
        CIFAR100 & 0.1 & 64.49 & 64.90 & 65.19 & 65.33 & 65.27 & \textbf{65.45} & 64.89 & 63.87 \\ 
        CIFAR100 & 0.2 & 63.71 & 63.74 & 63.89 & 64.05 & 64.24 & 64.23 & 64.10 & 64.30 \\ \hline
        STL10 & 0.1 & 89.34 & \textbf{89.94} & 89.87 & 89.84 & 89.72 & 89.52 & 88.99 & 88.41 \\
        STL10 & 0.2 & 88.4 & 88.23 & 88.4 & 88.35 & 87.54 & 88.32 & 88.80 & 88.59 \\ \hline
        Tiny-IN & 0.1 & 50.23 & 51.12 & 51.41 & 51.66 &  \textbf{51.90} & 51.58 & 51.37 & 50.46 \\
        Tiny-IN & 0.2 & 48.56 & 48.85 & 48.35 & 48.98 & 49.06 & 49.15 & 49.66 & 49.64 \\ \hline
    \end{tabular}
    \caption{Effect of varying the temperature parameters $\tau_m$ and $\tau$ on the Top-1 accuracy on small and medium datasets.}
    \label{tab:temperature-small}
\end{table*}

\section{Data augmentations details for evaluation protocol}


The data augmentations used for the evaluation protocol are:
\begin{itemize}[noitemsep,topsep=0pt]
    \item \textbf{training set for large datasets}: random crop to the resolution $224 \times 224$ and a random horizontal flip with a probability of $0.5$.
    \item \textbf{training set for small and medium datasets}: random crop to the dataset resolution with a padding of $4$ for small datasets and a random horizontal flip with a probability of $0.5$.
    \item \textbf{validation set for large datasets}: resize to the resolution $256 \times 256$ and center crop to the resolution $224 \times 224$.
     \item \textbf{validation set for small and medium datasets}: resize to the dataset resolution.
\end{itemize}

\section{Implementation details for pretraining small and medium datasets}

\textbf{Implementation details for small and medium datasets.} We use the ResNet-18 encoder and pretrain for $200$ epochs. Because the images are smaller, and ResNet is suitable for larger images, typically $224 \times 224$, we follow guidance from SimCLR and replace the first $7 \times 7$ Conv of stride $2$ with a $3\times3$ Conv of stride $1$. We also remove the first pooling layer. The strong data augmentation distribution applied is: random resized crop, color distortion with a strength of $0.5$, gray scale with a probability of $0.2$, gaussian blur with probability of $0.5$, and horizontal flip with probability of $0.5$. The weak data augmentation distribution is composed of a random resized crop and a random horizontal flip with the same parameters as the strong data augmentation distribution. 

We use 2 GPUs for a total batch size of 256. The memory buffer size is set to 4,096 for small datasets and 16,384 for medium datasets. The projector is a 2 fully connected layer network with a hidden dimension of 512 and an output dimension of 256. A batch normalization is applied after the hidden layer. The SGD optimizer is used during training with a momentum of $0.9$ and a weight decay of $5e^{-4}$. A linear warmup is applied during 5 epochs to reach the initial learning rate of $0.06$. The learning rate is scaled using the linear scaling rule: $lr = initial\_learning\_rate * batch\_size / 256$ and then follows the cosine decay scheduler without restart. The momentum value to update the momentum network is $0.99$ for small datasets and $0.996$ for medium datasets.

\section{Temperature influence on small and medium datasets}

We made a temperature search on CIFAR10, CIFAR100, STL10 and Tiny-ImageNet by varying $\tau$ in $\{0.1, 0.2\}$ and $\tau_m$ in $\{0.03, ..., 0.10\}$. The results are in \cref{tab:temperature-small}. As for ImageNet100, we need a sharper distribution on the output of the momentum encoder. Unlike ReSSL \cite{Zheng2021}, SCE do not collapse when $\tau_m \rightarrow \tau$ thanks to the contrastive aspect. For our baselines comparison in Sec. 4.2, we use the best temperatures found for each dataset.